# Optimal Design of Ad Hoc Injection Networks by Using Genetic Algorithms

Gregoire Danoy, Pascal Bouvry, Matthias R. Brust
University of Luxembourg
firstname.name@uni.lu

Enrique Alba
Department of Computer Science
University of Malaga
Malaga, Spain
eat@lcc.uma.es

## ABSTRACT

This work aims at optimizing *injection networks*, which consist in adding a set of long-range links (called *bypass links*) in mobile multi-hop ad hoc networks so as to improve connectivity and overcome network partitioning. To this end, we rely on small-world network properties, that comprise a high clustering coefficient and a low characteristic path length. We investigate the use of two genetic algorithms (generational and steady-state) to optimize three instances of this topology control problem and present results that show initial evidence of their capacity to solve it.

**Categories and Subject Descriptors:** I.2.8 [Computing Methodologies]: Artificial Intelligence: Problem Solving, Control Methods, and Search

## 1. INTRODUCTION

Mobile multi-hop ad hoc networks most often face the problem of network partitioning. In this work we consider the problem of optimizing *injection networks* which consist in adding of long-range links (e.g. GSM, UMTS or HSDPA) that are also called *bypass links* to interconnect network partitions. To tackle this topology control problem, we use small-world properties as indicators for the good set of rules to maximize the *bypass links* efficiency. Small-world networks [2] feature a high clustering coefficient ($\gamma$) while still retaining a small characteristic path length (L). One the one hand, a low characteristic path length is of importance for effective routing mechanisms as well as for the overall communication performance of the entire network. One the other hand, a high clustering coefficient features a high connectivity in the neighborhood of each node and thus a high degree of information dissemination each single node can achieve. This finally motivates the objective of evoking small-world properties in such settings. In order to optimize those parameters (maximizing $\gamma$, minimizing L) and to minimize the number of required bypass links in the network, we relied on Evolutionary Algorithms (EAs) and more specifically on Genetic Algorithms (GAs) [1]. We start by investigating the kind of evolution step more amenable to our problem by comparing both generational and steady-state GAs on a basic instance of a partitioned ad hoc network. At the same time, we investigate the influence of the crossover operators on the final quality of the results.

## 2. EXPERIMENTATIONS

Experiments have been conducted on three network scenarios, each scenario representing a snapshot of a mobile network in which nodes move in such a way that partitions are created. The three networks are composed of 30, 42 and 70 stations having respectively 5, 3 and 1 partitions. To simulate such hybrid networks, we used $M^AD^HO^C$, a network simulator that we extended in order to make it support bypass links and measure small-world properties. Every combination of GAs (generational and steady-state) and crossover operators (2-point and uniform) has been tested on these three network scenarios. The best results experimentally found on each of the three scenarios were obtained with a ssGA applying a 2-point crossover.

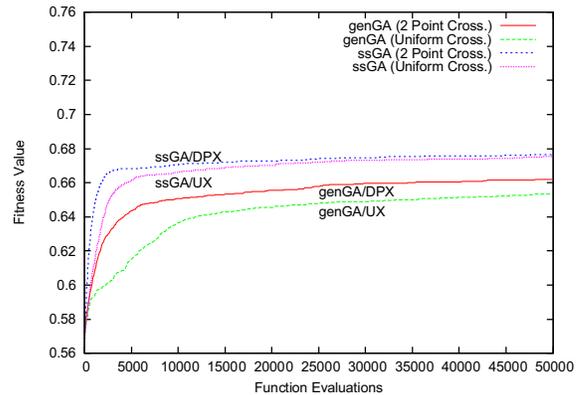

**Figure 1: Average results of 30 runs on the 3-partitions networks**

As a future work, we plan to use some more recent and effective GAs such as coevolutionary GAs to solve this problem. Our next research will also focus on the optimization of dynamic injection networks.